\documentclass[11pt]{article}

\usepackage[preprint]{acl}

\usepackage{times}
\usepackage{latexsym}

\usepackage[T1]{fontenc}
\usepackage[utf8]{inputenc}

\usepackage{microtype}

\usepackage{inconsolata}

\usepackage{amsmath,amssymb,amsfonts}
\usepackage{algorithmic}
\usepackage{algorithm}
\usepackage{graphicx}
\usepackage{textcomp}
\usepackage{xcolor}

\usepackage{xspace}

\usepackage{enumitem}
\usepackage{array}
\usepackage{booktabs}
\usepackage{multirow}
\usepackage{colortbl}
\usepackage{makecell}
\usepackage{adjustbox}
\usepackage{tcolorbox}
\usepackage{float}
\usepackage{xcolor}
\usepackage{balance}
\title{Knowledge Distillation for Low-Resource Open-source Text-to-SQL Model}

\tcbuselibrary{listings}  % If you need code highlighting support
\tcbuselibrary{listingsutf8} % If you need to support UTF-8 encoded code

\usepackage[T1]{fontenc}
\usepackage{geometry}
\usepackage{xcolor}
\usepackage{listings}

\definecolor{backcolour}{rgb}{0.95,0.95,0.92}
\definecolor{promptBlue}{rgb}{0.16, 0.32, 0.75}
\definecolor{promptGreen}{rgb}{0,0.5,0}
\definecolor{promptPurple}{rgb}{0.58,0,0.82}
\definecolor{promptOrange}{rgb}{1.0, 0.5, 0.0}
\definecolor{promptGray}{rgb}{0.5,0.5,0.5}

\lstdefinelanguage{Prompt}{
    backgroundcolor=\color{backcolour},
    basicstyle=\ttfamily\footnotesize,
    breaklines=true,
    breakatwhitespace=true,
    keepspaces=true,
    showstringspaces=false,
    tabsize=2,
    alsoletter={$,\texttt{\{}\,\texttt{\}}\,\_}, 
    keywords=[1]{
          \#\#\#, Domain, Knowledge, Relevant,pairs, Question
    },
    keywordstyle=[1]\color{promptBlue}\bfseries,
    keywords=[2]{Table,  continents, countries, ContId, Continent, CountryId, CountryName},
    keywordstyle=[2]\color{promptPurple},
    keywords=[3]{
       ${SCHEMA\_INFO},  ${DATABASE\_SCHEMA}, ${DATABASE\_CONTENT}, ${DOMAIN\_KG}, ${KG}, ${USER\_QUESTION},${QA\_PAIRS}, ${SCHEMA_EXPLANATION}
    },
    keywordstyle=[3]\color{promptOrange}\bfseries,
    keywords=[4]{
        Reasoning Component
    },
    keywordstyle=[4]\color{promptGreen},
    numbers=left,
    numberstyle=\tiny\color{promptGray},
    numbersep=5pt,
    commentstyle=\color{red},
}

\author{
  Tianhao Qiu \\
  Shenzhen University \\
  Shenzhen, China \\
  \texttt{2310275033@email.szu.edu.cn} \\
  \And
  Xiaojun Chen \\
  Shenzhen University \\
  Shenzhen, China \\
  \texttt{xjchen@szu.edu.cn} \\
}

\begin{document}
\maketitle
\begin{abstract}
Text-to-SQL converts natural language questions into executable SQL queries, enabling non-technical users to access relational databases for analytics and intelligent data services. In real-world scenarios, performance is often constrained by low-resource settings, where high-quality annotated \texttt{<question, SQL>} pairs are scarce, particularly for domain-specific databases. Additional challenges include opaque schema definitions, abbreviations, and implicit business logic that are not explicitly encoded in the schema. Existing data synthesis and prompting techniques improve coverage but often fail to produce task-specific, semantically grounded examples aligned with database constraints. To address these challenges, we propose a knowledge-aware Text-to-SQL framework that constructs task-specific knowledge base including schema semantics, abbreviations, business logic, and query patterns, and injects them into both training and inference. This framework generates diverse, contextually grounded synthetic training data and enhances inference through targeted knowledge retrieval. Experiments on seven benchmarks, covering both general and domain-specific datasets, demonstrate that our approach substantially improves the performance of open-source and closed-source large language models in Text-to-SQL tasks, especially in low-resource domain-specific settings, enhancing generalization, robustness, and adaptability.
\end{abstract}

\section{Introduction}
\label{sec:intro}

% 主要错误类型，逻辑错误/SCHEMA LINKING 错误 天乐  + 联系为啥要增强SCHEMA 类型 liu2025surveytexttosqlerallms
%  
    
Text-to-SQL is a foundational task in natural language processing that translates natural language questions into executable SQL queries. By serving as a bridge between non-technical users and relational databases, it enables intuitive and scalable access to structured data, powering applications such as business analytics, intelligent data services, and reporting. However, accurately mapping user intent to SQL while adhering to the strict syntactic and semantic constraints of relational schemas remains a core challenge~\cite{qin2022survey,geoyge2023survey}.

A key challenge in real-world Text-to-SQL is low-resource settings, where only a limited number of labeled \texttt{<question, SQL>} pairs are available for a given task—especially for open-source models, which cannot leverage proprietary data due to privacy constraints. Recent work has sought to mitigate this limitation through data synthesis strategies. Rule-based methods using grammars and templates~\cite{Yu2018@SyntaxSQLNet,Wu2021@sql2question-DA} and LLM-based approaches leveraging prompt engineering~\cite{li2024codes,yang2024synthesizing,li2025omnisqlsynthesizinghighqualitytexttosql} have expanded training coverage. However, these approaches primarily generate general-purpose samples and often fail to produce task-specific, semantically grounded examples, resulting in poor alignment with real-world database constraints~\cite{pourreza2023dinsql,wang2022proton}. 

To address this, we propose a framework to distill structured, task-specific knowledge from closed-source LLMs into open-source models. The framework captures domain terminology and SQL query patterns that encode semantic relationships between questions and schema elements. This knowledge is then leveraged to synthesize high-quality, grounded training examples for fine-tuning and to provide task-relevant context during inference, enhancing the model’s reasoning over complex queries. By transferring the implicit understanding of closed-source models to open-source ones, our framework enables more accurate, context-aware, and executable SQL generation, even in low-resource settings.

Our main contributions are as follows:
\begin{enumerate}[noitemsep, topsep=0pt, leftmargin=*, label=\arabic*.]
\item \textbf{Structured Knowledge Construction:} We develop a systematic approach to construct task-specific knowledge, including schema knowledge, domain terminology, and SQL query patterns. This includes algorithms for extracting domain terms and building the SQL Pattern Graph, which captures recurring relationships between question types and SQL skeletons.

\item \textbf{Knowledge-Aware Training and Inference:} Leveraging the constructed knowledge, we synthesize diverse and semantically accurate \texttt{<question, SQL>} pairs with LLMs for fine-tuning, and retrieve relevant schema, domain, and query pattern knowledge at inference to guide reasoning. This unified approach improves generalization, context-awareness, and the accuracy of SQL generation in low-resource and domain-specific settings.

\item \textbf{Extensive Evaluation:} We conduct comprehensive experiments across seven benchmarks spanning general and domain-specific datasets. Results demonstrate that our framework consistently enhances the performance of both open-source and closed-source LLMs, highlighting the value of structured knowledge for generalization, interpretability, and adaptability in real-world Text-to-SQL tasks.
\end{enumerate}

\section{Related Work}
\label{sec:related}

%  1. 

\subsection{Text-to-SQL}

Early Text-to-SQL solutions were mainly rule- or template-driven~\cite{li2014constructing,mahmud2015rule}, relying on handcrafted rules or SQL templates to convert natural language into queries. While effective for simple scenarios, they struggled to scale to complex, multi-domain settings due to rigidity and labor-intensive template design. Benchmark datasets such as WikiSQL~\cite{wikisql}, Spider~\cite{yu2018spider}, KaggleDBQA~\cite{kaggledbqa}, and BIRD~\cite{DBLP:journals/corr/abs-2305-03111} later enabled more realistic, multi-table, and cross-domain research.

With the rise of deep learning, Text-to-SQL has been reframed as a sequence-to-sequence problem. Encoder-decoder architectures~\cite{DBLP:conf/ijcai/CaiXZYLL18, DBLP:conf/coling/PopescuMVYKS22, qi2022rasat}, enhanced by attention mechanisms~\cite{DBLP:conf/ijcnn/LiuSZWLK23}, graph-based schema representations~\cite{DBLP:conf/emnlp/XuWWFS18, li2023graphix, DBLP:conf/acl/ZhengWDWL22, rat-sql}, and syntax-aware decoding~\cite{DBLP:conf/acl/GuoZGXLLZ19, scholak2021picard, li2023resdsql, wang2022proton}, have become dominant. Tabular Language Models like TaBERT~\cite{yin2020tabert} further support joint modeling of text and schema. Despite these advances, training such models remains expensive, and domain adaptation is challenging.

%Recently, large language models (LLMs) such as GPT~\cite{chatgpt,gpt4} and LLaMA~\cite{llama,llama2} have shown impressive Text-to-SQL capabilities. Three main paradigms have emerged: supervised fine-tuning~\cite{sun2023sql}, in-context learning (ICL)~\cite{icl22, nan2023enhancing, DBLP:journals/corr/abs-2303-13547, gao2023empowered}, and reinforcement learning (RL)~\cite{shao2024deepseekmath, pourreza2025reasoning, ma2025sql}. SFT updates model parameters using labeled pairs, ICL relies on prompts without modifying parameters, and RL leverages feedback to optimize behavior directly, improving robustness and alignment with complex objectives.

Recently, large language models (LLMs) such as GPT~\cite{chatgpt,gpt4} and LLaMA~\cite{llama,llama2} have demonstrated remarkable Text-to-SQL capabilities. Three main paradigms have emerged: \textbf{supervised fine-tuning (SFT)}~\cite{sun2023sql}, which updates model parameters using labeled \texttt{<question, SQL>} pairs; \textbf{in-context learning (ICL)}~\cite{icl22, nan2023enhancing, DBLP:journals/corr/abs-2303-13547, gao2023empowered}, which relies on carefully designed prompts without modifying model parameters; and \textbf{reinforcement learning (RL)}~\cite{shao2024deepseekmath, pourreza2025reasoning, ma2025sql}, which leverages feedback to directly optimize model behavior, improving robustness and alignment with complex objectives.

\begin{figure}[t]
    \centering
    \includegraphics[width=0.5\textwidth]{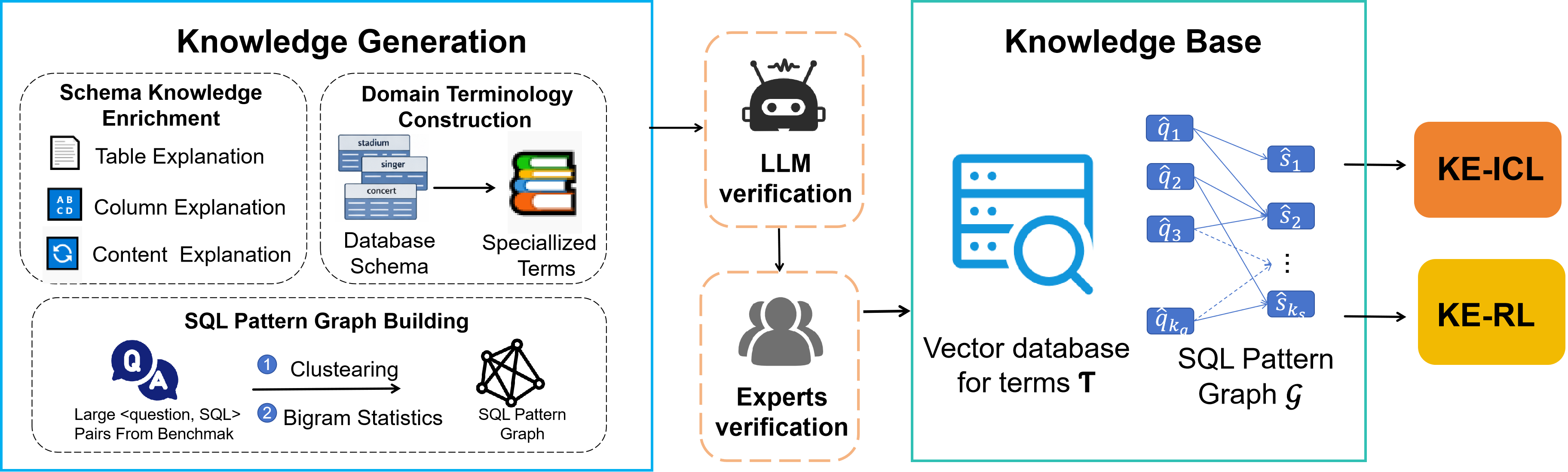}
    \caption{Our proposed knowledge enhancement framework for Text-to-SQL tasks.}
    \label{fig:framework}
\end{figure}

\subsection{Data Synthesis}

%A persistent challenge in Text-to-SQL is the scarcity of high-quality labeled data, especially for complex, multi-table, or domain-specific tasks. Annotating natural language and SQL pairs is expensive and often requires expert involvement, limiting model performance and generalization in low-resource settings.

Data synthesis methods aim to automatically generate additional \texttt{<question, SQL>} pairs to enhance training coverage, query diversity, and model robustness. Early rule-based approaches include template-driven generation, which translates manually crafted or database-derived SQL templates into questions~\cite{Guo2018@STAMP, Hu2023@importance-DA, li2024codes}; grammar-based generation, which constructs SQL via ASTs or grammars and converts them to natural language~\cite{Wu2021@sql2question-DA, Wang2021@learning-DA, Zhang2023@sciencebenchmark}; slot-filling, which populates reusable templates with schema elements or values~\cite{Yu2018@SyntaxSQLNet, Weir2020@dbpal, Yu2021@grappa, li2024codes}, though often producing repetitive or unnatural phrasing; and question-to-SQL with existing models, which generates questions first and predicts SQL, potentially introducing noise~\cite{Yang2021@hierarchical-DA}. While rule-based methods ensure structural correctness, they often struggle with scalability and semantic diversity. LLM-based synthesis has increasingly leveraged in-context learning with SQL templates, control prompts, and curated examples. For instance, Pourreza et al.~\cite{pourreza2024sql} select SQL templates from Spider to guide generation, Yang et al.~\cite{yang2024synthesizing} control SQL difficulty via table counts, and Li et al.~\cite{li2025omnisqlsynthesizinghighqualitytexttosql} synthesize diverse databases and systematically generate QA pairs with controlled complexity and language styles. However, because LLMs generate examples based on general patterns learned from broad pretraining data, they can lack domain grounding for a specific target database. This may result in syntactic or semantic mismatches with the schema and poor alignment with real-world constraints~\cite{pourreza2023dinsql,wang2022proton}.

\section{Motivation}

\label{sec:motivation}

External knowledge is crucial in complex Text-to-SQL tasks, helping models interpret user intent, align with database schemas, and generate valid queries. We categorize such knowledge into three types: 
1) \textbf{Schema Knowledge:} database structure, including table/column names, value formats, and relationships, enabling accurate schema linking;  
2) \textbf{Domain Knowledge:} task-specific concepts, terminology, and computation logic, allowing reasoning over derived metrics or expressions;  
3) \textbf{SQL Query Pattern Graph:} a structured representation of canonical SQL templates that capture typical reasoning patterns, modeling the mapping from question intent to SQL logic, including constructs such as subqueries, joins, and aggregations.

Closed-source LLMs may inherently capture portions of this knowledge, whereas open-source models often lack it entirely. To address this, we propose a unified knowledge distillation framework (Figure~\ref{fig:framework}) that constructs, verifies, and applies knowledge through a three-stage pipeline. Raw knowledge extracted from schema documentation, question–SQL pairs, and query clusters is first filtered and normalized using a lightweight LLM module combined with expert cross-validation. The verified knowledge is organized into four unidirectional tables, including $\mathcal{T}$ for domain terms and the SQL query pattern graph $\mathcal{G}$. This knowledge base enables: (i) \textbf{Knowledge-Enhanced In-Context Learning (KE-ICL)}, which enriches prompts and reduces ambiguity, and (ii) \textbf{Knowledge-Enhanced Reinforcement Learning (KE-RL)}, which generates diverse, schema-faithful training data to improve model robustness.

\section{Knowledge Construction}
\label{sec:kg_construction}

%   to do 改为多个模型生成知识，多个模型交叉检测
%   强调写出知识蒸馏的想法
%   推理   Token / 时间开销（时间开销） 

Our knowledge construction framework for Text-to-SQL consists of four stages: (i) \textbf{Schema Knowledge Enrichment}, which enhances schema semantics through clarified names and descriptions; (ii) \textbf{Domain Terminology Construction}, which maps domain-specific terms to SQL logic; (iii) \textbf{SQL Query Pattern Graph Building}, which builds a graph of query skeleton patterns; and (iv) \textbf{Knowledge Post-processing}, which validates and organizes the knowledge for context-aware SQL generation. The details of each stage are described below.

\subsection{Schema Knowledge Enrichment}

Schema knowledge can be enriched through a combination of domain experts and large language models (LLMs). The LLM infers metadata beyond the original schema definitions, generating human-readable annotations for table and column names, clarifying abbreviations, and interpreting encoded values to enhance semantic transparency. For instance, in the \texttt{california\_schools} database, a table \texttt{frpm} may be annotated as ``Free and Reduced-Price Meal Program statistics,'' \texttt{capacity} as ``number of available seats,'' and \texttt{dob} as ``date of birth.'' Similarly, value-level mappings such as \texttt{M}/\texttt{F} in \texttt{gender\_code} are interpreted as ``Male'' and ``Female.'' Domain experts can further validate and refine these annotations to ensure accuracy and consistency. By bridging low-level schema structures with natural language understanding, this enriched schema knowledge improves the model’s comprehension of database semantics, enabling more accurate and context-aware SQL generation.

\subsection{Domain Terminology Construction}

This stage constructs domain-specific terminology from database columns, as detailed in Algorithm~\ref{alg:dt_construction} in Appendix~\ref{app:dt_construction}. Each column is first encoded into a semantic embedding and clustered into groups representing related concepts. Candidate terms are then generated by sampling one term from each of two clusters and combining them with a sampled operator or symbol (\(op\)). Each candidate is validated by a large language model (LLM), which provides a validity label, confidence score, and an optional natural-language explanation. Valid terms are collected, and the top \(K\) terms are selected based on their confidence scores. This approach efficiently explores the space of column combinations while ensuring semantic diversity, interpretability, and high-quality domain terminology.

\subsection{SQL Pattern Graph Building}
\begin{figure}[t]
    \centering
    \includegraphics[width=0.5\textwidth]{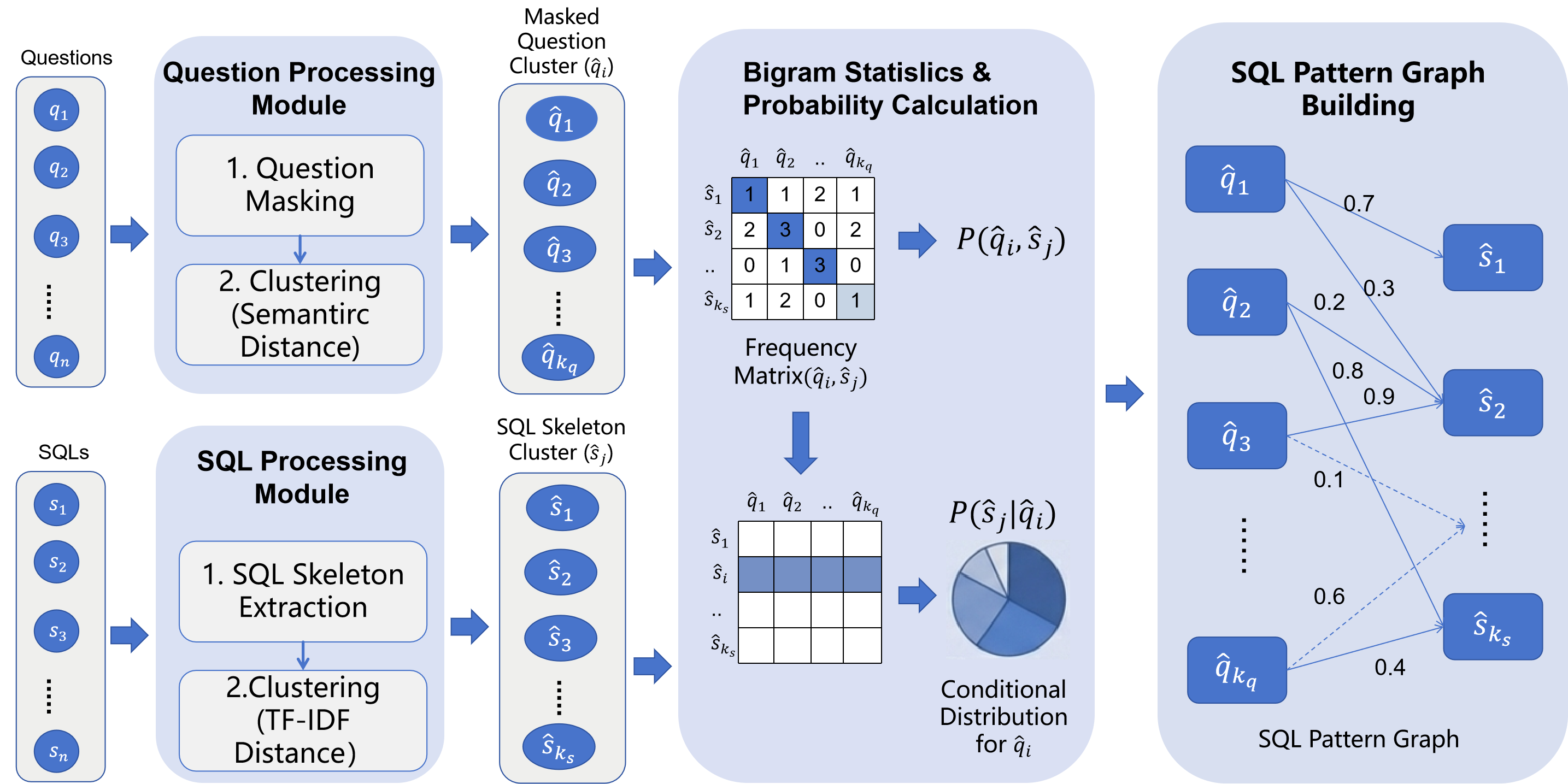}
    \caption{SQL Pattern Graph Construction}
    \label{fig:pattern_graph}
\end{figure}

The \textbf{SQL Pattern Graph} captures frequent mappings between question clusters and SQL skeleton clusters (Figure~\ref{fig:pattern_graph}). Let the input sets be questions $\mathcal{Q} = \{q_1, q_2, \dots, q_N\}$ and SQL answers $\mathcal{S} = \{s_1, s_2, \dots, s_N\}$. In the \textbf{Question Processing Module}, each question $q_i$ is first masked and then represented in a semantic space. Questions are clustered based on semantic distance, resulting in a set of $k_q$ question clusters $\{\hat{q}_{1},\cdots,\hat{q}_{k_q}\}$, where each cluster represents a type of skeleton. Simultaneously, in the \textbf{SQL Processing Module}, SQL skeletons are extracted from the SQL answers and clustered using TF-IDF features, results in a set of $k_s$ skeleton clusters $\{\hat{s}_{1},\cdots,\hat{s}_{k_s}\}$.

Next, \textbf{bigram statistics} between question clusters and SQL skeleton clusters are computed to estimate co-occurrence the frequencies $f(\hat{q}_i, \hat{s}_j)$ in the original question-sql pairs. From these, \textbf{conditional probabilities} are derived as $p(\hat{s}_j \mid \hat{q}_i) = \frac{f(\hat{q}_i, \hat{s}_j)}{\sum_{j'} f(\hat{q}_i, \hat{s}_{j'})}$. Finally, the \textbf{SQL Pattern Graph} $G = (\mathcal{V}, \mathcal{E})$ is constructed, where vertices $\mathcal{V} = \{\hat{q}_i, \hat{s}_j\}$ represent clusters and edges $e_{ij} \in \mathcal{E}$ connect question clusters to SQL skeleton clusters with weights corresponding to the conditional probabilities $w(e_{ij}) = p(\hat{s}_j \mid \hat{q}_i)$. This graph effectively encodes recurring patterns between questions and SQL templates, enabling structured generalization for SQL query generation.

\subsection{Knowledge Post-processing}
\label{sec:kg_po}

Before storage, all extracted knowledge undergoes a two-stage validation pipeline to ensure both semantic fidelity and execution reliability. \textbf{Schema knowledge} and \textbf{domain terminology} are evaluated on a per-database basis, reflecting their database-specific semantics, while \textbf{SQL query patterns} are validated globally across databases. Domain terminology and SQL patterns are assessed using a hybrid LLM–human framework with unified scoring criteria. Two state-of-the-art LLMs—Claude~3.5~Sonnet and Gemini~1.5~Pro—independently score each entry on a 1–5 scale along two dimensions: \textbf{semantic consistency} (whether the natural-language description faithfully captures the SQL logic) and \textbf{SQL validity} (whether the SQL is syntactically correct and executable). Entries receiving scores $\geq 4$ from both LLMs proceed to human validation, where two annotators independently apply the same criteria. Items with mutual agreement are accepted, while disagreements are resolved by a third expert adjudicator. This process ensures high precision at scale; on the BIRD benchmark, we construct an average of 20 validated domain terms per database, while SQL query patterns are validated across the full corpus.

Once validated, the knowledge is stored according to its scope and intended usage. \textbf{Schema knowledge} and \textbf{domain terminology} are maintained on a per-database basis: schema knowledge augments table and column metadata, while domain terminology is indexed in a database-specific vector store to enable precise semantic retrieval. On the BIRD benchmark, we construct an average of 20 validated domain terms for each database. In contrast, \textbf{SQL query patterns} are organized into a global \emph{SQL Query Pattern Graph} $\mathcal{G}$ shared across databases, capturing reusable reasoning structures that are independent of specific schema details. This graph is extracted from BIRD and OmniSQL~\cite{li2025omnisqlsynthesizinghighqualitytexttosql} and consists of approximately 50 question clusters and 150 SQL skeleton clusters, with edges encoding conditional associations between question intents and SQL structures. The graph is stored in a graph database (e.g., Neo4j) and retrieved during both training and inference to guide structured, cross-database SQL reasoning.

\section{Knowledge-Enhanced In-Context Learning}
\label{sec:knowledge_icl}

We propose a \textbf{knowledge-enhanced in-context learning (KE-ICL)} approach that constructs a composite prompt integrating structural, semantic, and contextual cues. The prompt follows a unified template (Listing~\ref{lst:sql_cot_prompt}), where the user’s question is inserted verbatim as \texttt{\$\{USER\_QUESTION\}}, and three additional components—\texttt{\$\{DATABASE\_SCHEMA\}}, \texttt{\$\{DOMAIN\_TERM\}}, and \texttt{\$\{QUERY\_PATTERN\}}—provide structured guidance.

\noindent\textbf{Database Schema and Domain Terms (\texttt{\$\{DATABASE\_SCHEMA\}}, \texttt{\$\{DOMAIN\_TERM\}}).}  
Both components leverage a single classifier, \textit{Knowledge Linker}, to predict the relevance of schema elements and domain-specific terms with respect to the user question. The classifier is trained on the BIRD dataset~\cite{li2024can} with a RoBERTa encoder~\cite{liu2019robert}, and outputs relevance scores for tables, columns, and domain terms. We select the top-$k_1$ tables and top-$k_2$ columns for schema linking, and the top-$k_3$ terms for domain knowledge.  

%To improve schema-query alignment, the filtered schema is enriched with semantically relevant database values using a BM25 index. Up to 4-gram phrases from the user question are matched to literal values in the database (e.g., ``Locally funded'' matched to \texttt{schools.FundingType}) and incorporated into the prompt, with a cap of 25 values per high-cardinality column. Domain terms are paired with their descriptions from \texttt{DK\_L} and included in the \texttt{\$\{DOMAIN\_TERM\}} component.

\noindent\textbf{SQL Query Pattern (\texttt{\$\{QUERY\_PATTERN\}}).} To facilitate analogical reasoning and pattern transfer, we retrieve the top-$k_4$ SQL skeletons from the SQL query pattern graph $\mathcal{G}$. Given a query $q$, we first identify the two most similar question clusters, $\hat{q}_1$ and $\hat{q}_2$, and compute the conditional probability of each SQL skeleton cluster $\hat{s}_j$ as$
p(\hat{s}_j \mid q) = p(\hat{q}_1) \cdot p(\hat{s}_j \mid \hat{q}_1) + p(\hat{q}_2) \cdot p(\hat{s}_j \mid \hat{q}_2)$, 
where $p(\hat{q}_i)$ is proportional to the similarity between $q$ and $\hat{q}_i$. The top-$k_4$ skeletons are then sampled based on these probabilities using weighted random sampling, selecting the most relevant patterns to include as in-context examples for guiding SQL generation.

\begin{figure}[t]
    \centering
    \includegraphics[width=0.5\textwidth]{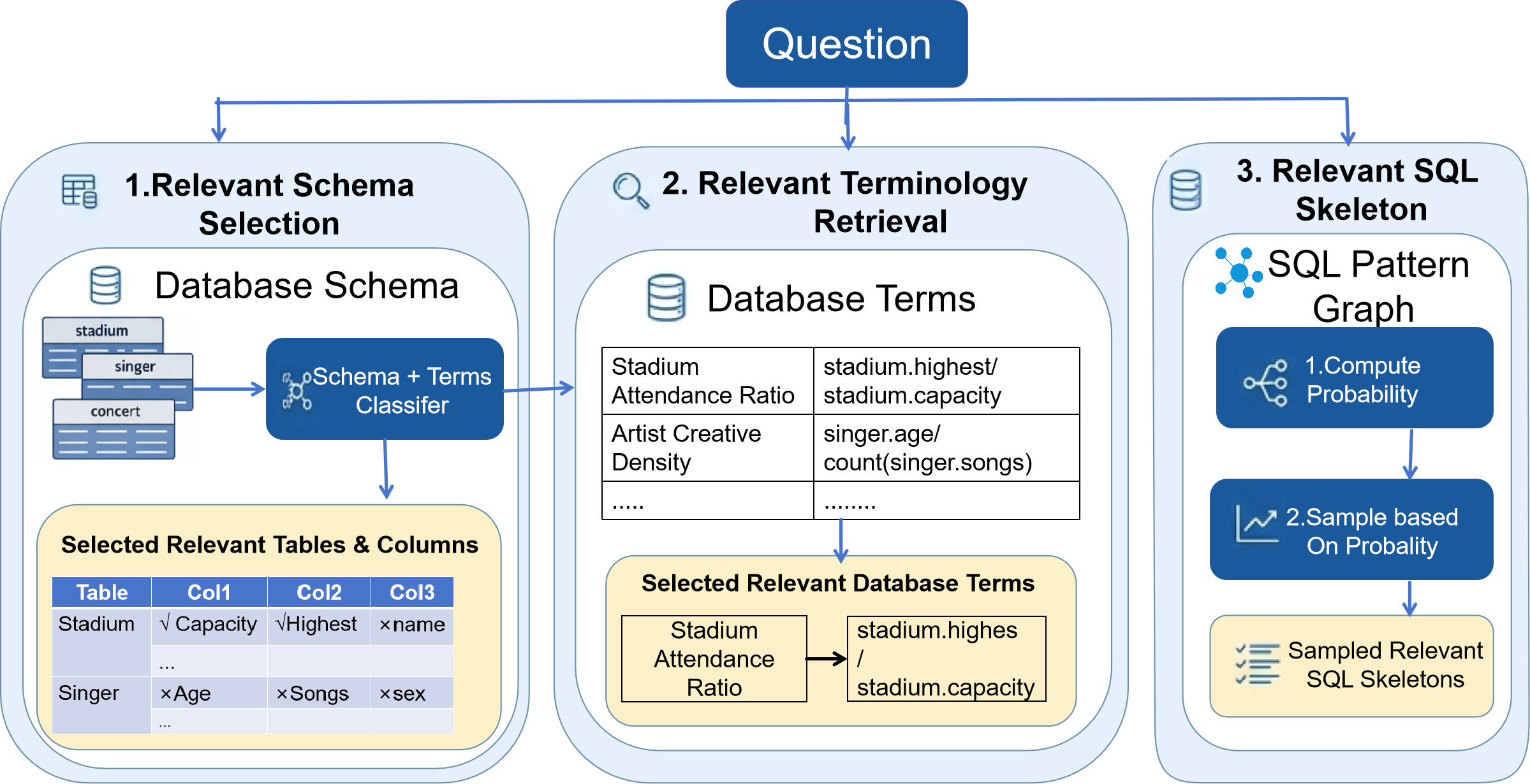}
    \caption{\label{fig:ga_gen} Knowledge-Enhanced In-Context Learning.}
\end{figure}

\section{Knowledge-Enhanced Reinforcement Learning}
\label{sec:knowledge_rl}

Existing open-source LLMs often struggle with Text-to-SQL tasks, particularly in incorporating domain-specific knowledge, limiting performance on specialized databases. We propose \textbf{Knowledge-Enhanced Reinforcement Learning (KE-RL)}, which leverages schema and domain knowledge to generate diverse, accurate, and contextually grounded question–SQL pairs for LLM training. As shown in Figure~\ref{fig:ga_gen}, the pipeline consists of four stages: SQL template generation, knowledge-aware Q–SQL pair generation, data augmentation, and GRPO-based fine-tuning.

\noindent\textbf{SQL Template Generation: }We construct abstract SQL skeletons capturing high-level query structure (e.g., \texttt{SELECT}, \texttt{WHERE}, \texttt{GROUP BY}) while masking schema-specific elements. The skeleton pool combines 440 extracted and 100 manually designed skeletons, covering common and complex patterns. Each skeleton is expanded into executable SQL templates using LLM-guided synthesis with three difficulty levels: Easy, Medium, Hard. Templates are retained adaptively based on the number of tables involved, yielding 3,854 validated SQL templates.

\noindent\textbf{Knowledge-Aware Question--SQL Pair Generation}
Templates are sampled based on structural diversity and similarity to representative queries $\mathcal{R}$ from the SQL query pattern graph $\mathcal{G}$:
\[
N = \left\lceil \frac{\rho}{1-\rho} \times \frac{M}{16} \right\rceil, \quad
p_i = 
\begin{cases}
\frac{S_i^\alpha}{\sum_j S_j^\alpha}, & T\neq 0\\
\frac{1}{N}, & T=0
\end{cases}
\]
where $S_i$ is the average cosine similarity between template $t_i$ and representative queries, $\alpha$ controls the sampling bias ($\alpha > 0$ favors templates similar to known queries, $\alpha = 0$ yields uniform sampling, and $\alpha < 0$ encourages selection of structurally diverse templates)
, and $\rho$ determines the synthetic-to-real data ratio.  

For each sampled template, a knowledge-enriched prompt incorporates both the database schema and the top-$k$ relevant domain knowledge entries. An LLM generates the SQL statement and corresponding natural language question, which are validated for syntactic correctness, schema consistency, and semantic alignment.

\noindent\textbf{Data Augmentation: }To enhance diversity, each validated pair undergoes SQL rewriting (three semantically equivalent variants per original SQL) and question rephrasing (three alternative phrasings per SQL variant), producing 16 Q–SQL pairs per example. Chain-of-thought explanations are minimally adapted to reflect edits.

\noindent\textbf{GRPO Training: }The synthesized and augmented dataset is used to fine-tune the LLM with GRPO, guided by a knowledge-aware, execution-driven reward:

\begin{equation*}
\footnotesize
R_i =
\begin{cases}
1, & \text{if the SQL execution matches the ground truth;} \\
0.5, & \text{if the SQL respects knowledge constraints;} \\
0.1, & \text{if the SQL is executable;} \\
0, & \text{otherwise.}
\end{cases}
\end{equation*}
The intermediate reward for partial alignment encourages the model to respect relevant knowledge—such as schema and domain-term constraints—even if the execution result is not fully correct. This knowledge-aware reward guides GRPO to generate SQL queries that are syntactically valid, semantically accurate, and consistent with both the database schema and associated domain knowledge, promoting robust reasoning over structured data.
\section{Experiment}
\label{sec:exp}

\subsection{Main Result}

\begin{table*}[t]
\centering
\caption{Execution accuracy on seven benchmarks.}
\label{tab:codes_vs_us}
\begin{adjustbox}{width=\textwidth}
\renewcommand{\arraystretch}{1.1}
\setlength{\tabcolsep}{3pt}
\begin{tabular}{l|l|ccc|ccc|cc|c}
\toprule
\multirow{2}{*}{LLM} & \multirow{2}{*}{Method} 
& \multicolumn{3}{c|}{Standard} 
& \multicolumn{3}{c|}{Robustness} 
& \multicolumn{2}{c|}{Domain-Specific} 
& \multirow{2}{*}{Average} \\
\cmidrule(lr){3-5} \cmidrule(lr){6-8} \cmidrule(lr){9-10}
& & Spider-dev & Spider-test & BIRD-dev & Spider-DK & Spider-Syn & Spider-Realistic & EHRSQL & Science Benchmark & \\
\midrule

\multirow{3}{*}{GPT-4o}
 & DAIL-SQL (ICL) & 72.45 & 85.03 & 62.18 & 64.91 & 63.44 & 75.51 & 40.02 & 53.62 & 64.65 \\
 & CodeS (ICL) & 73.88 & 86.16 & 64.41 & 65.98 & 65.86 & 76.77 & 41.27 & 54.84 & 66.15 \\
 & \textbf{KE-ICL} & \textbf{74.17} & \textbf{86.40} & \textbf{65.25} & \textbf{68.41} & \textbf{66.05} & \textbf{78.14} & \textbf{47.12} & \textbf{59.53} & \textbf{68.13} \\
\hline

\multirow{3}{*}{Gemini-Pro-1.5} 
 & DAIL-SQL (ICL) & 78.91 & 87.20 & 66.30 & 73.02 & 69.50 & 78.30 & 51.00 & 50.20 & 69.30 \\
 & CodeS (ICL) & 80.17 & 88.31 & 67.54 & 74.20 & 71.17 & 79.72 & 52.28 & 51.54 & 70.62 \\
 & \textbf{KE-ICL} & \textbf{80.27} & \textbf{88.68} & \textbf{67.80} & \textbf{75.32} & \textbf{71.76} & \textbf{80.11} & \textbf{55.46} & \textbf{53.17} & \textbf{71.57} \\
\hline

\multirow{6}{*}{Deepseek-Coder-7B}  
 & DAIL-SQL (ICL) & 61.44 & 70.90 & 38.90 & 63.95 & 53.13 & 56.02 & 13.41 & 28.88 & 48.33 \\
 & CodeS (ICL) & 63.53 & 72.38 & 40.29 & 65.34 & 55.18 & 57.13 & 14.88 & 29.76 & 49.81 \\
 & \textbf{KE-ICL} & \textbf{67.89} & \textbf{75.08} & \textbf{45.82} & \textbf{68.10} & \textbf{64.00} & \textbf{66.20} & \textbf{23.51} & \textbf{34.44} & \textbf{55.62} \\
 & SQL-GEN (RL) & 74.05 & 79.73 & 49.87 & 70.30 & 59.65 & 62.72 & 28.30 & 34.62 & 57.41 \\
 & Omni (RL) & 78.90 & 82.78 & 52.70 & 72.47 & 62.27 & 64.85 & 31.46 & 39.29 & 60.47 \\
 & \textbf{KE-RL} & \textbf{80.49} & \textbf{84.76} & \textbf{58.74} & \textbf{76.58} & \textbf{66.04} & \textbf{68.73} & \textbf{45.64} & \textbf{48.32} & \textbf{66.16} \\
\hline

\multirow{6}{*}{Granite-3.1-8B}  
 & DAIL-SQL (ICL) & 57.01 & 67.20 & 35.44 & 46.92 & 43.01 & 46.10 & 12.30 & 31.89 & 42.48 \\
 & CodeS (ICL) & 59.15 & 68.56 & 37.16 & 48.17 & 44.63 & 47.43 & 13.59 & 33.11 & 43.98 \\
 & \textbf{KE-ICL} & \textbf{61.22} & \textbf{70.24} & \textbf{41.78} & \textbf{52.27} & \textbf{47.78} & \textbf{50.13} & \textbf{20.63} & \textbf{36.12} & \textbf{47.52} \\
 & SQL-GEN (RL) & 69.32 & 77.91 & 52.23 & 60.12 & 55.44 & 57.01 & 33.10 & 46.88 & 56.50 \\
 & Omni (RL) & 77.14 & 82.34 & 49.98 & 56.97 & 52.21 & 55.10 & 30.70 & 41.13 & 55.69 \\
 & \textbf{KE-RL} & \textbf{78.01} & \textbf{82.48} & \textbf{57.48} & \textbf{62.71} & \textbf{58.09} & \textbf{60.93} & \textbf{41.48} & \textbf{49.16} & \textbf{61.29} \\
\hline

\multirow{6}{*}{Qwen2.5-Coder-7B}  
 & DAIL-SQL (ICL) & 74.05 & 84.44 & 54.70 & 70.43 & 65.43 & 63.71 & 19.84 & 38.22 & 58.85 \\
 & CodeS (ICL) & 75.73 & 85.61 & 56.13 & 71.29 & 64.77 & 64.80 & 21.03 & 39.79 & 59.89 \\
 & \textbf{KE-ICL} & \textbf{77.85} & \textbf{85.98} & 56.06 & \textbf{74.02} & 66.98 & \textbf{71.45} & 33.43 & 43.14 & 63.61 \\
 & SQL-GEN (RL) & 77.56 & 85.32 & 57.92 & 73.90 & 67.80 & 70.35 & 34.51 & 44.12 & 63.94 \\
 & Omni (RL) & 78.43 & 85.14 & 58.60 & 76.52 & 69.17 & 72.43 & 38.00 & 46.48 & 65.60 \\
 & \textbf{KE-RL} & \textbf{82.34} & \textbf{87.97} & \textbf{65.82} & \textbf{80.97} & \textbf{74.83} & \textbf{77.58} & \textbf{47.93} & \textbf{56.50} & \textbf{71.74} \\

\bottomrule
\end{tabular}
\end{adjustbox}
\end{table*}

Experiment details are provided in Appendix~\ref{app:experiment_setup}. As shown in Table~\ref{tab:codes_vs_us}, our knowledge-enhanced methods, KE-ICL and KE-RL, consistently deliver substantial gains across all benchmarks, LLMs, and average metrics. KE-ICL achieves the highest average performance among ICL methods, surpassing the second-best baseline (CodeS) by +3.2\% across five LLMs, with notable improvements on Deepseek-Coder-7B (+5.81\%) and Granite-3.1-8B (+3.54\%), and smaller yet consistent gains on GPT-4o (+1.98\%), Gemini-Pro-1.5 (+0.95\%), and Qwen2.5-Coder-7B-Instruct (+3.72\%). These results highlight the effectiveness of leveraging schema components, domain-specific expressions, and representative query exemplars for inference-time reasoning. KE-RL further strengthens performance, outperforming the strongest RL baseline (Omni) by +5.8\% on average across open-source LLMs, with particularly large gains on Deepseek-Coder-7B (+16.35\% over CodeS), Granite-3.1-8B (+5.60\%), and Qwen2.5-Coder-7B-Instruct (+6.14\%). Beyond standard benchmarks, KE-RL enhances robustness against paraphrasing and ambiguity, and delivers remarkable domain-specific improvements—+26.90\% on EHRSQL and +16.71\% on ScienceBenchmark compared to CodeS. These findings demonstrate that training with structured, knowledge-informed synthetic data effectively improves syntactic validity, semantic accuracy, and reasoning over both schema and domain knowledge, providing a scalable and cost-efficient pathway for building high-performing Text-to-SQL systems with open-source LLMs.

\begin{table*}[t]
\centering
\caption{Ablation study of knowledge-enhanced prompting with KE-SI on Qwen2.5-Coder-7B-Instruct. ``↓'' indicates performance drop compared to the full setting, with absolute deltas in parentheses. \textbf{Bold} highlights the most affected metrics per row.}
\label{tab:ablation_qwen}
\begin{adjustbox}{width=\textwidth}
\renewcommand{\arraystretch}{1.2}
\setlength{\tabcolsep}{4pt}
% New column specification with vertical separators for benchmark groups
\begin{tabular}{l|ccc|ccc|cc|c}
\toprule
\multirow{2}{*}{\textbf{Knowledge Setting}} 
& \multicolumn{3}{c|}{\textbf{Standard}} 
& \multicolumn{3}{c|}{\textbf{Robustness}} 
& \multicolumn{2}{c|}{\textbf{Domain-Specific}} 
& \multirow{2}{*}{\textbf{Average}} \\
\cmidrule(lr){2-4} \cmidrule(lr){5-7} \cmidrule(lr){8-9}
& \textbf{Spider-dev} & \textbf{Spider-test} & \textbf{BIRD (dev)} & \textbf{Spider-DK} & \textbf{Spider-Syn} & \textbf{Spider-Realistic} & \textbf{EHRSQL} & \makecell{\textbf{Science}\\\textbf{Benchmark}} & \\
\midrule
\midrule
ALL Knowledge 
& 82.34 & 87.97 & 65.82 & 80.9 & 74.83 & 75.58 & 47.93 & 56.50 & 71.74 \\
\midrule
w/o Enhanced Schema Info   
& \textbf{75.58 ↓ (6.76)} & \textbf{82.80 ↓ (5.17)}  & \textbf{58.98 ↓ (6.84)} & \textbf{72.32 ↓ (8.65)} & \textbf{68.23 ↓ (6.60)} & \textbf{67.21 ↓ (8.37)} & 43.80 ↓ (4.13) & \textbf{51.60 ↓ (4.90)} & \textbf{65.07 ↓ (6.68)} \\

w/o Representative Queries 
& 79.83 ↓ (2.51) & 86.29 ↓ (1.68) & 59.50 ↓ (6.32) & 76.50 ↓ (4.47) & 68.24 ↓ (4.59) & 72.66 ↓ (4.92) & \textbf{41.10 ↓ (6.83)} & 52.40 ↓ (4.10) & 67.07 ↓ (4.68) \\

w/o Domain Terminology     
& 78.48 ↓ (3.86) & 85.31 ↓ (2.66) & 61.62 ↓ (4.10) & 74.22 ↓ (6.75) & 70.75 ↓ (4.08) & 69.63 ↓ (5.95) & 45.10 ↓ (2.83) & 53.30 ↓ (3.20) & 67.30 ↓ (4.4) \\
\bottomrule
\end{tabular}
\end{adjustbox}
\end{table*}

% \begin{table}[ht]
% \centering
% \caption{Test On BIRD Dev}
% \label{tab:codes_vs_us}
% \begin{tabular}{l r}
% \hline
% \textbf{Method} & \textbf{EX(DEV)} \\
% \hline
% AskData + GPT-4o\cite{wu2022dataaugmentationhierarchicalsqltoquestion} & 75.36 \\
% CHASE-SQL Gemini \cite{Pourreza2024@chasesql} & 74.90 \\
% XiYan-SQL\cite{Gao2024@xiyansql} & 73.34 \\
% OmniSQL-32B\cite{li2025omnisqlsynthesizinghighqualitytexttosql} & 69.23 \\
% MCS-SQL\cite{lee2024mcssqlleveragingmultipleprompts} & 63.36 \\
% GPT-4o & 57.95 \\
% DAIL-SQL + GPT-4 \cite{Gao2024@dailsql} & 54.76 \\
% TA-SQL + GPT-4 \cite{qu2024tasql} & 56.19 \\
% SFT CodeS-15B \cite{li2024codes} & 58.47 \\
% DTS-SQL + DeepSeek 7B \cite{Pourreza2024@dtssql} & 55.80 \\
% KE-RL QWEN2.5-Coder 7B & 60.62 \\
% \hline
% \end{tabular}
% \end{table}

% \begin{table}[ht]
% \centering
% \caption{Test on Spider}
% \label{tab:spider_res}
% \begin{tabular}{l r}
% \hline
% \textbf{Method} & \textbf{test} \\
% \hline
% MCS-SQL + GPT-4  \cite{lee2024mcssqlleveragingmultipleprompts}& 89.60 \\
% CHASE-SQL + Gemini \cite{Pourreza2024@chasesql}1.5 & 87.60 \\
% DAIL-SQL + GPT-4 \cite{Gao2024@dailsql}& 86.60 \\
% DIN-SQL + GPT-4 \cite{Pourreza2023@dinsql} & 85.30 \\
% GPT-4o & 83.54 \\
% OmniSQL-32B \cite{li2025omnisqlsynthesizinghighqualitytexttosql}& 89.80 \\
% C3 + ChatGPT + Zero-Shot \cite{dong2023c3} & 82.30 \\
% DTS-SQL + DeepSeek 7B \cite{Pourreza2024@dtssql} & 84.40 \\
% KE-RL QWEN2.5-Coder 7B & 86.87 \\
% \hline
% \end{tabular}
% \end{table}
\subsection{Schema Linking Result}

% ~\cite{chen2024opensql}

We evaluate our proposed two-step schema linking strategy (Section~\ref{sec:knowledge_icl}) using Schema Linking Recall (SLR)\cite{maamari2024deathsl}, which measures the proportion of questions for which all required columns are correctly retrieved—a prerequisite for accurate SQL generation. As shown in Table~\ref{tab:schema_linking}, two main findings emerge. First, the Step 1 schema classifier consistently improves over the LLM-only baseline across different numbers of retained columns ($k_2$). For example, on BIRD-dev, increasing $k_2$ from 2 to 8 gradually improves SLR, demonstrating that structure-aware filtering effectively ranks the most relevant columns. Even at lower $k_2$ values, Step 1 outperforms the LLM, highlighting the schema classifier’s superior ability to select relevant tables and columns. Second, Step 2, which adds term expansion via value-aware retrieval, further increases recall over Step 1 across most $k_2$ settings. Gains are most pronounced at moderate $k_2$ values, while at very high $k_2$, improvements are smaller, suggesting that Step 1 already ranks relevant columns effectively. Overall, our two-step schema linking method consistently outperforms the LLM, providing robust schema linking and enhanced coverage for downstream SQL generation.

\begin{table}[t]
\centering
\caption{Effect of varying the top-$k_3$ retrieved knowledge items on model performance(execution accuracy) on the BIRD-DEV benchmark across different LLMs.}
\label{tab:birddev_knowledge_injection}
\resizebox{0.5\textwidth}{!}{  % 控制表格宽度为页面的一半
\begin{tabular}{|l|c|c|c|c|}
\hline
\textbf{Top-$k_3$} & \makecell[c]{\textbf{Qwen-2.5}\\\textbf{Coder-7B}} & \textbf{GPT-4o-mini} & \textbf{GPT-4o} & \textbf{Gemini-1.5} \\
\hline
\textbf{$k_3$=0}     & 54.70 & 61.83 & 64.41 & 67.54 \\
\hline
\textbf{$k_3$=3}       & 55.10 & 62.51 & 64.50 & 67.30 \\
\hline
\textbf{$k_3$=5}       & 56.06 & 63.34 & 65.25 & 67.80 \\
\hline
\textbf{$k_3$=7}       & 54.90 & 64.36 & 65.28 & 68.22 \\
\hline
\textbf{$k_3$=9}       & 54.20 & 62.56 & 64.43 & 66.36 \\
\hline
\end{tabular}
}  % 结束 resizebox
\end{table}

\subsection{In-Context Learning}

\noindent\textbf{Impact of Knowledge Types.} Following the setup in Appendix~\ref{app:experiment_setup}, we assess the contribution of different knowledge types under the KE-SI setting using the Qwen2.5-Coder-7B-Instruct model. As shown in Table~\ref{tab:ablation_qwen}, each injected component—Enhanced Schema Info, Representative Queries, and Domain Terminology—plays a distinct and complementary role in enhancing in-context learning. Enhanced Schema Info proves to be the most critical. Its removal leads to the largest performance declines, particularly on Spider-DK ($-8.65\%$), Spider-Realistic ($-8.37\%$), and ScienceBenchmark ($-4.90\%$), underscoring its importance in establishing structural alignment for accurate SQL generation. Representative Queries also contribute substantially, especially in domain specific benchmarks such as EHRSQL ($-6.83\%$) and ScienceBenchmark ($-4.10\%$). While domain terminology has a relatively smaller impact, it remains important for grounding semantic understanding in real applications. Its removal causes notable drops on Spider-Realistic ($-5.95\%$) and ScienceBenchmark ($-3.20\%$), highlighting its role in entity disambiguation and domain-specific reasoning. Overall, the removal of any individual component results in consistent performance degradation across benchmarks, emphasizing the necessity of holistic knowledge injection for robust in-context learning.

\noindent\textbf{Impact of Knowledge Quantity.} In this experiment, we examine the impact of injecting domain knowledge (DK) entries and representative questions (RQ) by varying $k_3$, while keeping the number of selected tables fixed at $k_1 = 5$ and the number of selected columns per table fixed at $k_2 = 12$. As shown in Table~\ref{tab:birddev_knowledge_injection}, three key findings emerge. 
%  错误修改  knowledge consistently improves performance compared with the $k_3=0$ baseline 
First, adding knowledge consistently improves performance compared with the $k_3=0$ baseline, validating its usefulness for semantic grounding. For example, GPT-4o improves from 64.41\% at $k_3=0$ to 65.25\% at $k_3=5$. Second, performance gains peak at moderate levels ($k_3=5$ or $7$), whereas larger injections ($k_3=9$) often degrade accuracy—indicating that excessive context may introduce noise. Third, the impact varies across models: GPT-4o and Gemini-1.5 achieve their best results at $k_3=7$, while Qwen-2.5-Coder-7B shows only marginal improvements. Overall, controlled knowledge injection enhances robustness, but excessive information can hinder performance.

% We evaluate how varying the number of injected Domain Knowledge (DK) and Representative Questions (RQ), which was controled by paramter $k_3$ (See Section~\ref{sec:knowledge_icl}) affects model performance on BIRD-DEV. As shown in Table~\ref{tab:birddev_knowledge_injection}, injecting a moderate quantity of knowledge (Top-5) leads to consistent gains across all evaluated models. For example, Qwen2.5-Coder-7B-Instruct improves from 54.70\% (w/o DK RQ) to 56.06\%, and GPT-4o from 64.41\% to 65.25\%. However, increasing the number to Top-7 leads to mixed effects: while Gemini-1.5 continues to benefit, reaching its peak score of 68.22\%, smaller models like Qwen2.5-Coder-7B-Instruct (54.90\%) and GPT-4o-mini (64.36\%) show signs of stagnation or regression. With Top-9 DK, RQ, performance slightly drops or plateaus across most models, suggesting that excessive context can lead to redundancy or interfere with the model’s reasoning. These results emphasize the need to balance information quantity in prompt construction, tailoring injection volume to model capacity for optimal performance.

\subsection{Reinforcement Learning}
\begin{figure}[t]
    \centering
    \includegraphics[width=0.48\textwidth]{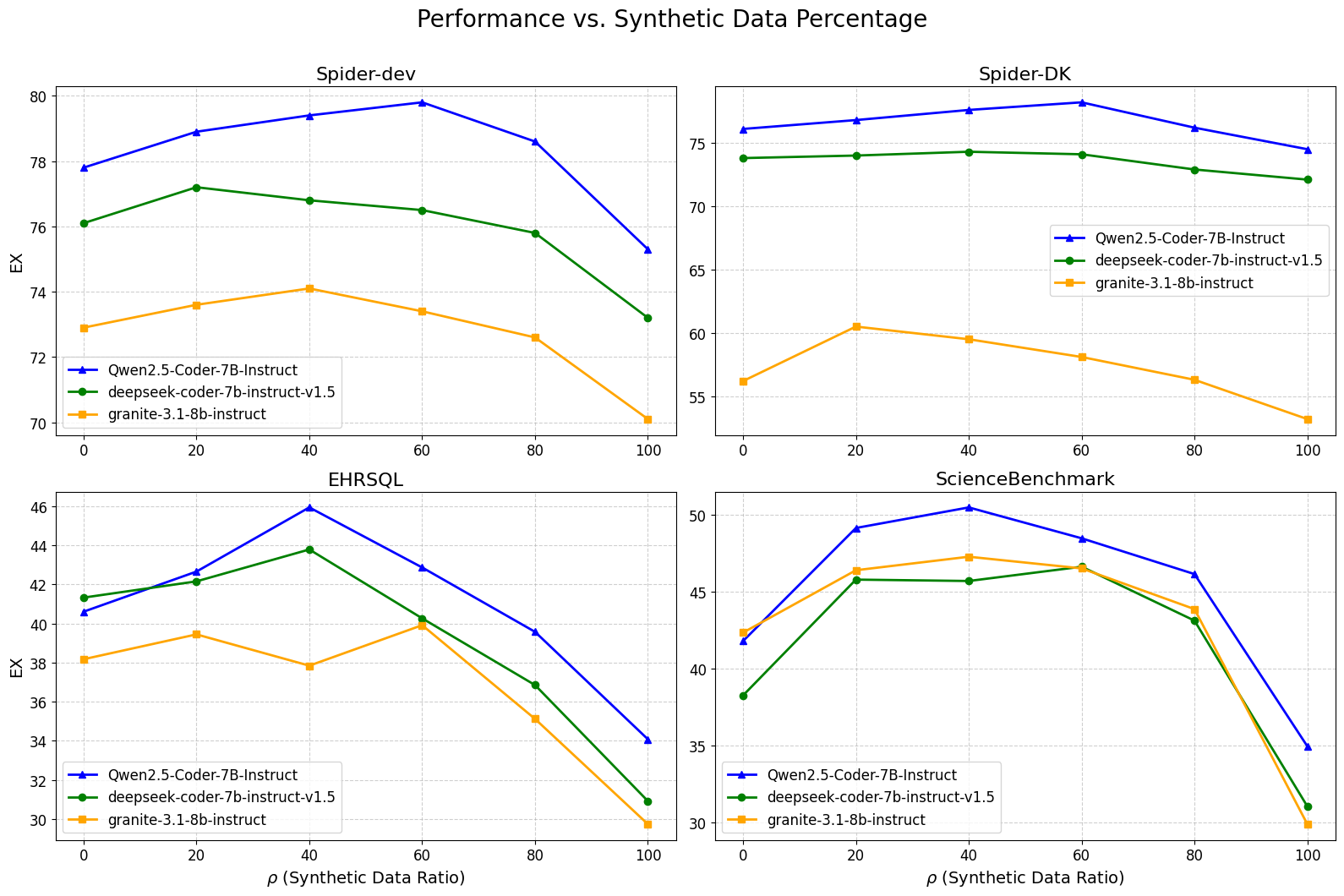}
    \caption{Impact of varying synthetic data ratio $\rho$ on execution accuracy, with total training size fixed at 5,000.}
    \label{fig:alb_rate}
\end{figure}

\noindent\textbf{Effect of Synthetic Data Ratio.} The synthetic-to-real data ratio $\rho$ controls the proportion of generated samples relative to human-annotated ones. 
We study how varying $\rho$ influences model performance while keeping the 
total training size fixed at 5,000 instances. As shown in Figure~\ref{fig:alb_rate}, performance exhibits an inverted U-shaped trend: mixing real and synthetic data consistently outperforms using either alone. For \textbf{domain-specific benchmarks} (EHRSQL and ScienceBenchmark), 
optimal performance arises when 20–40\% of the data is synthetic. 
In this range, synthetic data broadens coverage while real annotations 
anchor domain-specific logic. In contrast, for \textbf{standard and robustness benchmarks} 
(Spider-dev and Spider-DK), higher synthetic ratios (40–80\%) are more effective, 
as synthetic data enriches query diversity and strengthens generalization to unseen structures. When $\rho = 1.0$, no real data remains, and template sampling reduces to uniform selection across all templates without representative query guidance, disrupting query pattern preferences and leading to sharp performance drops across all datasets. Overall, these results underscore a key trade-off: \emph{with limited annotation budgets, 
neither relying solely on synthetic data nor exclusively on real data is optimal. 
Instead, carefully tuning $\rho$ enables practitioners to maximize performance under 
realistic resource constraints}.

\begin{figure}[t]
    \centering
    \includegraphics[width=0.48\textwidth]{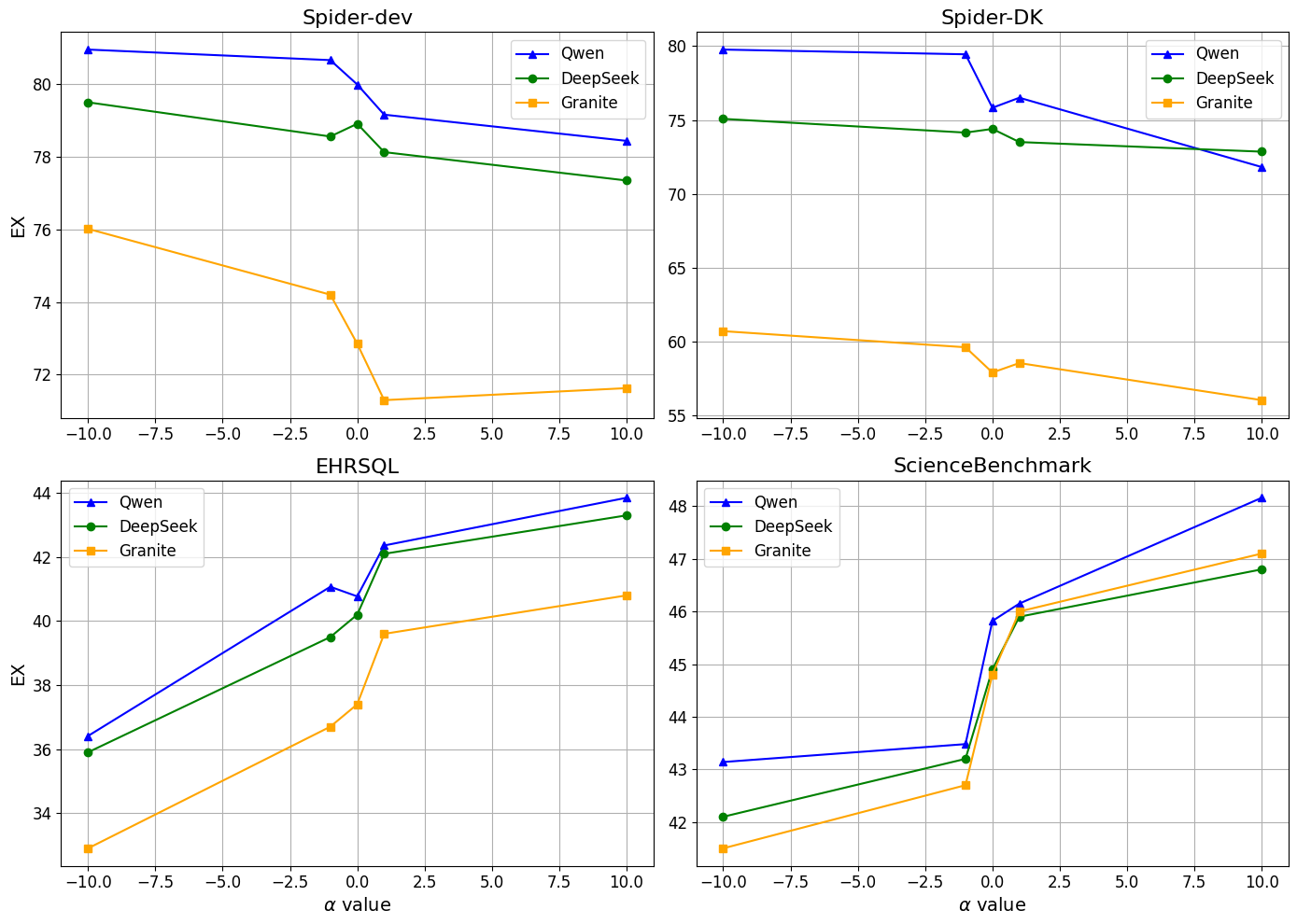}
    \caption{Effect of the sampling bias parameter $\alpha$ on model execution accuracy.}
    \label{fig:alb_alpha}
\end{figure}

\noindent\textbf{Effect of Template Sampling Strategy in Data Synthesis} In our data synthesis pipeline, the hyperparameter $\alpha$ controls the balance between favoring query patterns in the training set and exploring novel query patterns. As shown in Figure~\ref{fig:alb_alpha}, the effect of $\alpha$ is task-dependent. 
For \textbf{domain-specific benchmarks} (EHRSQL and ScienceBenchmark), accuracy 
improves steadily as $\alpha$ increases, peaking at $\alpha = 10$. 
This suggests that emphasizing templates closer to known query patterns 
is essential for capturing domain-specific logic. In contrast, for \textbf{standard and robustness benchmarks} (Spider-dev and Spider-DK), performance is highest when $\alpha < 0$, indicating that encouraging structural diversity 
enhances cross-domain generalization. 
Overall, these results reveal a trade-off between \textbf{relevance-driven sampling}, 
which strengthens specialization, and \textbf{diversity-driven sampling}, 
which improves robustness. This underscores the importance of adapting $\alpha$ 
to different benchmark settings.

\section{Conclusion}
\label{sec:conclusion}
We propose a unified knowledge-aware Text-to-SQL framework that incorporates task-specific domain knowledge—including schema semantics, abbreviations, and business logic—into both training and inference. By generating diverse, contextually grounded synthetic data and performing targeted knowledge retrieval over SQL skeletons and representative query patterns, the framework enhances reasoning and semantic grounding, addressing challenges of data scarcity and structural complexity. Experiments on seven benchmarks demonstrate substantial performance gains across open- and closed-source LLMs, improving generalization, interpretability, and adaptability. Future work will focus on more advanced knowledge retrieval strategies that reason over skeleton structures and multi-hop dependencies.

\newpage

\section{Limitations}
\label{sec:limitation}
Despite its effectiveness, the framework has several limitations. First, building and maintaining a high-quality knowledge base requires significant effort and domain expertise, which can limit scalability and increase costs, particularly for large or frequently evolving databases. Second, outputs generated by LLMs may still exhibit hallucinations or inconsistencies with domain constraints, which can reduce reliability. Third, databases with very large schemas or complex table relationships can strain knowledge retrieval and reduce the effectiveness of template coverage, limiting scalability and efficiency.
\bibliography{custom}
% Custom bibliography entries only
% \bibliographystyle{acl_natbib.bst}
% \bibliography{custom}

\onecolumn
\appendix

\section{Experiment Setup}
\label{app:experiment_setup}

\noindent\textbf{Benchmarks.} We utilized three distinct benchmark sets to assess our proposed method. (1) \textbf{Standard Benchmarks:} We use the BIRD dataset~\cite{DBLP:journals/corr/abs-2305-03111} (BIRD-dev split, 1,534 examples) and Spider~\cite{yu2018spider} (Spider-dev: 1,034; Spider-test: 2,147). These benchmarks evaluate structured query generation over cross-domain databases. (2) \textbf{Robustness Benchmarks:} We employ Spider-DK~\cite{deng2020spiderdk}, Spider-Syn~\cite{gan2020spidersyn}, and Spider-Realistic\allowbreak~\cite{li2020spiderrealistic} to assess model robustness. These datasets cover domain-specific reasoning, column name paraphrasing, and realistic query variations, containing 535, 1,034, and 508 queries, respectively. (3) \textbf{Domain-Specific Benchmarks:} We adopt EHRSQL~\cite{wan2023ehrsql} and ScienceBenchmark~\cite{sciencebenchmark2023} to evaluate performance in specialized domains. EHRSQL consists of 1,008 clinical queries, while ScienceBenchmark includes 299 queries across disciplines such as policy, astronomy, and oncology.

\noindent\textbf{Baselines.} We compare our approach with a diverse set of models and enhancement strategies. For \textbf{ICL-based baselines}, we evaluate Knowledge-Enhanced In-Context Learning (KE-ICL) against prompt-based models such as DAIL-SQL~\cite{gao2023empowered} and CodeS~\cite{li2024codes}, optimized for lightweight inference via single-pass prompting. This evaluation covers both commercial and open-source Large Language Models (LLMs): GPT-4o~\cite{openai2024@gpt4o}, Gemini-Pro-1.5~\cite{team2024gemini}, Deepseek-Coder-7B-Instruct-v1.5~\cite{Guo2024@deepseek-coder}, Qwen2.5-Coder-7B-Instruct~\cite{hui2024@qwen2.5-coder}, and Granite-3.1-8B-Instruct~\cite{Mishra2024@granite-coder}. For \textbf{RL-based baselines}, we compare Knowledge-Enhanced Reinforcement Learning (KE-RL) with SQL-GEN~\cite{pourreza2024sql} and OmniSQL~\cite{li2025omnisqlsynthesizinghighqualitytexttosql}, two leading data synthesis frameworks for text-to-SQL learning. This comparison, as well as fine-tuning, is conducted exclusively on the open-source LLMs: Deepseek-Coder-7B-Instruct-v1.5, Qwen2.5-Coder-7B-Instruct, and Granite-3.1-8B-Instruct. All RL methods use the CodeS prompting strategy during training and inference.

\noindent\textbf{Metrics.} We use execution accuracy (EX)~\cite{DBLP:journals/corr/abs-2305-03111}, which measures the correctness of query results and enables fair, database-agnostic comparisons.

\noindent\textbf{Implementation Details.} All experiments are conducted on 8 NVIDIA A100 80GB GPUs.

\textit{Domain Terminology Construction.} In Domain Terminology Construction columns from the database,  are first encoded into semantic embeddings using \texttt{RoBERTa}. The columns are then clustered into groups based on semantic similarity, with the number of clusters \(M\) determined by the maximum silhouette coefficient. Candidate terms are generated by sampling one column from each of two clusters and combining them with a random operator. These candidate terms are validated by large language models, including \texttt{GPT-4.0 (GPT-4-turbo)}, \texttt{Claude 3.5}, \texttt{Deepseek v2}, and \texttt{Qwen 1.5}, with one model randomly chosen for each validation. The top \(K = 150\) terms are selected based on confidence scores, and the process continues until \(N_{\text{target}} = 300\) valid terms are generated.

\textit{Training Data Synthesis.} For standard benchmarks (Spider and BIRD) and their robustness variants, we fine-tune on the BIRD training set augmented via KE-RL, with a synthetic data ratio $\rho = 0.6$ and strong negative sampling bias ($\alpha = -10$) to encourage structurally diverse SQL templates. This yields $\sim$14,706 new question--SQL pairs, combined with 9,821 original examples for a total of $\sim$24,527. For domain-specific benchmarks (EHRSQL, ScienceBenchmark), we preprocess 5,000 examples (EHRSQL) and augment ScienceBenchmark from 300 to 5,000, then generate 7,500 synthetic examples each ($\rho = 0.6$, $\alpha = 10$), resulting in 12,500 examples per dataset. Baselines use similar-scale synthetic datasets (SQL-GEN and OmniSQL: 13,000 for BIRD, 7,500 for EHRSQL/ScienceBenchmark).

\textit{Reinforcement Learning.} We fine-tune using GRPO in a prompt-completion format. Prompts use knowledge injection with top-5 tables ($k_1=5$), top-6 columns per table ($k_2=6$), and top-3 relevant QA examples ($k_3=3$), truncating \texttt{\$\{DATABASE\_SCHEMA\}} to fit a 2,048-token context. GRPO settings: temperature 0.8, total batch 256 (16 rollouts), update batch 128, KL penalty $\beta=0.001$, clip ratio $\epsilon=0.2$. LoRA (rank $r=8$) is applied to $q_\text{proj}$, $k_\text{proj}$, $v_\text{proj}$ layers, with pretrained weights frozen. Training uses per-device batch size 1, gradient accumulation 2, bf16 precision, under DeepSpeed ZeRO-3~\cite{zero2020}.

\textit{Inference.} The same knowledge-injection strategy retrieves top-5 tables ($k_1=5$), top-12 columns per table ($k_2=12$), and top-5 QA examples ($k_3=5$), within 4,096 tokens. Self-consistent decoding generates 8 SQL candidates per question via nucleus sampling (top-$p=0.95$). Executable candidates are evaluated against the target database, selecting the query returning the most results, with ties broken by shortest execution time.

\section{Lack konwledge of wrong case}
\begin{table}[t]
\caption{
Illustrative examples of SQL correction by GPT-4o through the injection of three types of knowledge.}
\centering
\small
\renewcommand{\arraystretch}{1.2}
\begin{tabular}{p{\linewidth}}
\toprule
\textbf{Example 1: Schema Knowledge — Column Description} \\
\textbf{Question:} What's the reference name of Marina Bay Street Circuit? (BIRD-SQL) \\
\textbf{Incorrect SQL generated without external knowledge:} \texttt{SELECT name FROM circuits ...} \\
\textbf{Injected knowledge:} \texttt{``reference name'' $\rightarrow$ circuits.circuitRef} \\
\textbf{Corrected SQL generated with knowledge injection:} \texttt{SELECT circuitRef FROM circuits ...} \\
\midrule
\textbf{Example 2: Schema Knowledge — Abbreviations} \\
\textbf{Question:} What are the websites for all the partially virtual chartered schools in San Joaquin? (BIRD-SQL) \\
\textbf{Incorrect SQL generated without external knowledge:} \texttt{...WHERE virtual = 'partial'} \\
\textbf{Injected knowledge:} \texttt{``partially virtual'' $\rightarrow$ Virtual = 'P'} \\
\textbf{Corrected SQL generated with knowledge injection:} \texttt{...WHERE Virtual = 'P'} \\
\midrule
\textbf{Example 3: Domain Knowledge — Specialized Terms} \\
\textbf{Question:} What is the complete address of the school with the lowest excellence rate? (BIRD-SQL) \\
\textbf{Incorrect SQL generated without external knowledge:} \texttt{SELECT address ... ORDER BY excellence\_rate ASC} \\
\textbf{Injected knowledge:} \texttt{``complete address'' $\rightarrow$ CONCAT(...)} \\
\texttt{``excellence rate'' $\rightarrow$ NumGE1500 / NumTstTakr} \\
\textbf{Corrected SQL generated with knowledge injection:} \texttt{SELECT CONCAT(...) ... ORDER BY (NumGE1500 / NumTstTakr) ASC} \\
\midrule
\textbf{Example 4: Representative Queries — Relevant Query} \\
\textbf{Question:} What’s the cost to get glucocorticoids - methylprednisolone? (EHRSQL) \\
\textbf{Incorrect SQL generated without external knowledge:} \texttt{SELECT cost FROM treatment ...} \\
\textbf{Injected knowledge:} \\
\texttt{Q: how much does it cost for a hemothorax diagnosis?} \\
\texttt{A: select distinct cost.cost from cost where cost.eventtype = 'diagnosis' and cost.eventid in ....} \\
\textbf{Corrected SQL generated with knowledge injection:} \texttt{SELECT cost FROM cost WHERE eventid IN (SELECT treatmentid FROM treatment ...)} \\
\bottomrule
\end{tabular}

\label{tab:kg_examples}
\end{table}

\begin{table*}[t]
\centering
\caption{Schema Linking Recall (\%) results by Qwen2.5-Coder-7B with three schema linking approaches: the \textbf{LLM} method directly uses Qwen-2.5-Coder-7B-Instruction to identify required tables and columns without any schema-specific preprocessing, as in~\cite{chen2024opensql}; \textbf{Step 1} performs schema classifier training and relevant schema prediction; and \textbf{Step 2} further enhances Step 1 with term expansion via value-aware retrieval, as detailed in Section~\ref{sec:knowledge_icl}.}
\label{tab:schema_linking}
\resizebox{\textwidth}{!}{%
\begin{tabular}{@{}l c ccc ccc cc@{}}
\toprule
\multirow{2}{*}{\textbf{Method}} & 
\multirow{2}{*}{\makecell{\textbf{Retained Columns} \\ \textbf{per Table} ($k_2$)}} & 
\multicolumn{3}{c}{\textbf{Standard}} & 
\multicolumn{3}{c}{\textbf{Robustness}} & 
\multicolumn{2}{c}{\textbf{Domain-Specific}} \\
\cmidrule(lr){3-5} \cmidrule(lr){6-8} \cmidrule(lr){9-10}
& & Spider-dev & Spider-test & BIRD (dev) & Spider-DK & Spider-Syn & Spider-realistic & EHRSQL & ScienceBenchmark \\
\midrule
\textbf{LLM}
& -  & 90.61  & 91.29 & 85.79 & 89.53 & 88.20 & 87.20 & 96.03 & 90.63 \\
\midrule
\multirow{4}{*}{\textbf{Step 1 Only}} 
& 4  & 97.87 & 98.37 & 93.87 & 98.13 & 95.84 & 94.68 & 92.64 & 98.40 \\
& 8  & 99.41 & 99.61 & 98.04 & 100.0 & 99.41 & 99.21 & 97.99 & 99.30 \\
& 12 & 99.61 & 99.76 & 98.63 & 100.0 & 99.61 & 99.61 & 98.66 & 99.70 \\
& 16 & 99.61 & 99.76 & 99.08 & 100.0 & 99.61 & 99.61 & 98.66 & 99.70 \\
\midrule
\multirow{4}{*}{\textbf{Steps 1 \& 2}} 
& 4  & 97.87 & 98.45 & 95.76 & 98.13 & 95.93 & 94.88 & 93.31 & 99.10 \\
& 8  & 99.41 & 99.67 & 98.95 & 100.0 & 99.41 & 99.21 & 98.32 & 99.43 \\
& 12 & 99.61 & 99.80 & 99.34 & 100.0 & 99.61 & 99.61 & 99.00 & 99.70 \\
& 16 & 99.61 & 99.80 & 99.61 & 100.0 & 99.61 & 99.61 & 99.00 & 99.70 \\
\bottomrule
\end{tabular}%
}
\end{table*}

\newpage
\section{Domain Terminology Construction Algorithm}
\label{app:dt_construction}

\begin{algorithm}
\caption{Domain Terminology Construction with Term Combination}
\label{alg:dt_construction}
\begin{algorithmic}[1]
\STATE \textbf{Input:} Column set \(C\) from a database, number of clusters \(M\), top-\(K\), target number of terms \(N_{\text{target}}\)
\STATE \textbf{Output:} Top \(K\) validated terms with explanations

\STATE Compute embeddings \(\mathbf{e}_c = \text{Encoder}(c)\) for each \(c \in C\)
\STATE Cluster columns into semantic groups \(\mathcal{K} = \{K_1, \dots, K_M\}\)

\STATE Initialize candidate term set \(\mathcal{T} = \{(K_i, \text{null}) \mid K_i \in \mathcal{K}\}\)  

\STATE Initialize counter \(n_{\text{valid}} = 0\)

\WHILE{\(n_{\text{valid}} < N_{\text{target}}\)}
    \FOR{each pair of terms \((t_i, t_j) \in \mathcal{T}, i \neq j\)}
        \STATE Sample an operator (e.g., \(+,-,*,/\)) as \(op\)
        \STATE Form combined candidate term \(t = \text{Combine}(t_i, op, t_j)\)
        \STATE Validate \(t\) using LLM: \((y_t, s_t, e_t) = \text{LLM\_Review}(t)\)
        \IF{\(y_t\) is valid} 
            \STATE Add \((t, e_t)\) to \(\mathcal{T}\)
            \STATE Increment \(n_{\text{valid}} = n_{\text{valid}} + 1\)
        \ENDIF
    \ENDFOR
    \STATE Remove any candidate terms with fewer than 2 columns
\ENDWHILE
\STATE Rank \(\mathcal{T}\) by confidence \(s_t\) and return the top \(K\) terms with explanations
\end{algorithmic}
\end{algorithm}

\newpage
\section{Prompt for inference prompt}
\label{app:prompt_inference}

\begin{lstlisting}[language=Prompt, caption={Example of \texttt{SQL GENERATION COT ROMPT}}, float=t, label={lst:sql_cot_prompt}]
You are tasked with generating a SQL query according to a input user request.
Note that
1. If the column name contains special characters such as spaces, please use ` to enclose it.
2. Exactly select the columns that the user wants to select, and do not select other unnesssary columns.
3. Once you need to subquery, please use CTE that starts with the WITH keyword to wrap the subquery and give it a name.
4. The final Answer Query **must** be wrapped in Markdown format using triple backticks and the `sql` tag.
5. You must reason step by step using a compositional approach. Your reasoning process should follow a **minimal set of steps** selected from a predefined library of 10 reasoning components (listed below). 
### Reasoning Components (Choose From):
1. Intent Recognition  
2. Disambiguation  
3. Temporal Reasoning  
4. Keyword Mapping  
5. Constraint Extraction  
6. Aggregation & Grouping Reasoning  
7. Ordering & Limiting  
8. Alias & Expression Handling  
9. Join Reasoning  
10. Nested/Subquery Reasoning
### DATABASE SCHEMA
${DATABASE_SCHEMA}
### DOMAIN KNOWLEDGE 
${DOMAIN_KG}
### RELEVANT QA PAIRS 
${QA_PAIRS}
### QUESTION  
${USER_QUESTION}
Please think step by step: 
\end{lstlisting}

\newpage
\section{Cost Analysis}
\label{subsec:syn_ratio_exp}
% \begin{table}[ht]
% \centering
% \small % 缩小整个表格的字体
% \caption{Token and Time Cost Analysis Based on Column Count in the BIRD Benchmark using the Gemini 1.5 Pro model.}
% \label{tab:diff_col_db_cost}
% \resizebox{0.5\textwidth}{!}{  % 控制表格宽度为页面的一半
% \begin{tabular}{@{}l c c c c c@{}}
% \toprule
% Category & \makecell[c]{\#AVG.Tables} & \makecell[c]{\#AVG.Columns} & \makecell[c]{Existing \\ Queries} & \makecell[c]{Average \\ Token Cost (1k)} & \makecell[c]{Average \\ Time Cost (s)} \\
% \midrule
% Few Columns  & 4.0  & 15.7 & 87.4 &  1283.2  & 714.7 \\
% Medium Columns & 4.9  & 33.8  & 120.1 & 2168.6   & 1163.1 \\
% Many Columns  & 13.5 & 101.5 & 199.0  & 5676.3  & 4365.4 \\
% \bottomrule
% \end{tabular}
% } % 结束 resizebox
% \end{table}
\begin{table}[t]
\centering
\small % 缩小整个表格的字体
\caption{Token and Time Costs vs. Column Count for Data Generation Methods.}
\label{tab:diff_col_db_cost}
\resizebox{0.5\textwidth}{!}{% 控制表格宽度为页面的一半
\begin{tabular}{@{}c c c c c c c c@{}}
\toprule
% 使用 \multirow{2}{*}{...} 让前两个单元格跨越两行
% {2} 表示跨越的行数
% {*} 表示宽度由内容自动决定
% \makecell 用于更好地控制单元格内对齐和换行
\multirow{2}{*}{\makecell[c]{Category}} & \multirow{2}{*}{\makecell[c]{\#AVG.\\Columns}} & \multicolumn{3}{c}{\textbf{Token Cost (1k)}} & \multicolumn{3}{c}{\textbf{Time Cost (s)}} \\
\cmidrule(lr){3-5} \cmidrule(lr){6-8}
% 第二行表头，前两列是空的，因为被 \multirow 占据了
& & \makecell[c]{SQL-Gen} & \makecell[c]{Omini} & \makecell[c]{Ours} & \makecell[c]{SQL-Gen} & \makecell[c]{Omini} & \makecell[c]{Ours} \\
\midrule
\makecell[c]{Few Columns}  & 15.7 & 581.4 & 3428.3 & 1283.2 & 319.4 &  1182.3 & 714.7 \\
\makecell[c]{Medium Columns} & 33.8 & 1324.1 & 7421.6 & 2168.6 & 397.5 & 1524.6 & 1163.1 \\
\makecell[c]{Many Columns}  & 101.5 & 3823.7 & 19723.8 & 5676.3  & 531.9 & 1914.5 & 2365.4 \\
\bottomrule
\end{tabular}
} % 结束 resizebox
\end{table}

In this section, we analyze the token consumption and time costs of our knowledge-aware Text-to-SQL framework, which is essential for evaluating its scalability and efficiency, particularly given the reliance on Large Language Models (LLMs). We conducted an experiment using the Gemini 1.5 Pro model on the BIRD Benchmark training set. The databases were divided into three groups based on their number of columns, using quartiles to split the dataset into three equal parts: \textbf{Few columns} (\(< 24\) columns), \textbf{Medium columns} (\(24 \leq \text{columns} \leq 48\)), and \textbf{Many columns} (\(> 48\) columns). For each group, we randomly selected five databases and synthesized 300 samples per database, using the Gemini 1.5 Pro model for data generation.

Table~\ref{tab:diff_col_db_cost} compares token and time costs for SQL-Gen, Omini, and our method (Ours) on databases with varying column counts in the BIRD Benchmark. While SQL-Gen is the most efficient in terms of both token and time costs, its data quality is lower compared to Omini and our method, leading to poorer model performance. Omini incurs high token and time costs due to its self-consistency technique but generates better quality data. Our method has higher time costs, especially with larger databases, due to the knowledge synthesis step, which handles Schema Knowledge. Despite higher costs, both Omini and our method produce better results than SQL-Gen.

\begin{table}[t]
\centering
\caption{Token and Time Consumption for Different Phases in the Knowledge-Aware Data Synthesis on the BIRD Benchmark using the Gemini 1.5 Pro model.}
\label{tab:token_consumption}
\resizebox{0.5\textwidth}{!}{  % 控制表格宽度为页面的一半
\begin{tabular}{@{}l l c c c@{}}
\toprule
\multirow{1}{*}{Phase} & \multirow{1}{*}{Task Name} & Input Tokens & Output Tokens & Times \\ 
\midrule
\multirow{3}{*}{\makecell[l]{Knowledge \\Construction}} 
& Schema Knowledge & 413.2 & 44.9 & 709.7 \\ 
& Domain Knowledge & 283.8 & 65.8 & 370.6 \\ 
& Knowledge Checking & 73.5 & 12.6 & 76.5 \\ 
& Total & 770.5 & 123.3 & 1,158.8 \\ 
\midrule
\multirow{3}{*}{\makecell[l]{Data \\Synthesis}} 
& Question-SQL Gen & 296.3 & 97.5 & 195.9 \\ 
& Data Correction Checking & 368.2 & 17.5 & 76.5 \\ 
& Data Augment & 1168.3 & 209.1 & 650.5 \\ 
& Total & 1824.8 & 324.1 & 922.8 \\ 
\bottomrule
\end{tabular}
}
\end{table}

Table~\ref{tab:token_consumption} summarizes the token and time consumption across the Knowledge Construction and Data Synthesis phases of the knowledge-aware pipeline. The Knowledge Construction phase incurs relatively low computational costs, with the Schema Knowledge and Domain Knowledge tasks being the most resource-intensive. In contrast, the Data Synthesis phase, particularly the Data Augmentation step, exhibits much higher token usage (1,168.3 input tokens) and time cost (650.5 seconds), making it the main computational bottleneck. These results suggest that Knowledge Construction could be further optimized by leveraging pre-existing contextual knowledge to reduce training overhead, while improvements in Data Synthesis—especially the augmentation process—would yield the greatest efficiency gains. Overall, knowledge construction is relatively lightweight, whereas data augmentation remains the primary challenge for scalability and computational efficiency in knowledge-aware frameworks.
% This is an appendix.

\end{document}